\documentclass[letterpaper]{article} % DO NOT CHANGE THIS
\usepackage{aaai2027}  % DO NOT CHANGE THIS
\usepackage[hyphens]{url}  % DO NOT CHANGE THIS
\usepackage{graphicx} % DO NOT CHANGE THIS
\usepackage{natbib}  % DO NOT CHANGE THIS AND DO NOT ADD ANY OPTIONS TO IT
\usepackage{caption} % DO NOT CHANGE THIS AND DO NOT ADD ANY OPTIONS TO IT
\usepackage{algorithm}
\usepackage{algorithmic}
\usepackage{comment}
\usepackage{graphicx}
\usepackage{booktabs}
\usepackage{multirow}
\usepackage{soul}
\usepackage{xcolor}
\usepackage[table]{xcolor}   % 支持 \rowcolor，并启用 table 选项
\usepackage{subcaption}      % 支持 subtable 环境
\usepackage[accsupp]{axessibility}  % Improves PDF readability for those with disabilities.

\usepackage[table]{xcolor}
\usepackage{newfloat}
\usepackage{listings}
\DeclareCaptionStyle{ruled}{labelfont=normalfont,labelsep=colon,strut=off} % DO NOT CHANGE THIS
\floatstyle{ruled}
\newfloat{listing}{tb}{lst}{}
\floatname{listing}{Listing}

\title{Proxy Avatar Meets Low-Rank Caching: Real-Time One-Shot Emotion-Controllable Portrait Animation}

\author{
    Haijie Yang$^1$,
    Jindi Bao$^1$,
    Yixuan Dong$^2$,
    Hongliang Zhang$^1$,
    Jian Bi$^1$,
    Hao Tang$^3$,
    Zhenyu Zhang$^4$,
    Jianjun Qian$^1$, 
    Jian Yang$^1$
}
\affiliations{
    \textsuperscript{\rm 1}Nanjing University of Science and Technology\\
    \textsuperscript{\rm 2}Tsientang Institute for Advanced Study\\
    \textsuperscript{\rm 3}Peking University\\
    \textsuperscript{\rm 4}Nanjing University\\
}

\begin{document}

\maketitle

%Large Language Models (LLMs) have advanced Audio-Visual Speech Recognition (AVSR), yet they struggle in real-world scenarios due to their disregard for paralinguistic cues and reliance on static resource allocation. To address these challenges, \textbf{PUCA-AVSR} is introduced to substantially enhance the performance and efficiency of AVSR. Our approach includes three key components: A \textbf{multi-task paralinguistic encoder} that captures paralinguistic cues like emotion or prosody, helping the model resolve linguistic ambiguities and adapt to diverse speaking styles, thereby achieving higher accuracy in transcribing emotionally rich or expressively complex speech.
%A \textbf{self-supervised audio complexity predictor} that estimates the signal-level difficulty of processing the audio. Using a teacher-student training setup, this lightweight module generates frame-level complexity pseudo-labels from unlabeled audio, enabling efficient prediction of comprehension difficulty.
%A \textbf{dynamic modality gating mechanism} coupled with a complexity-guided resource allocation strategy which adaptively adjusts the fusion weights of audio and visual inputs according to signal quality, moving beyond static fusion methods.
%Extensive experiments demonstrate that PUCA-AVSR showcases superior accuracy, robustness, and computational efficiency across various datasets.

\begin{abstract}
%Audio-driven portrait animation has advanced rapidly with diffusion-based generative models, yet real-time one-shot generation with fine-grained emotion control remains challenging. Existing methods often suffer from weak emotion-aware motion priors and heavy appearance computation during multi-step denoising. To address these issues, we propose Proxy Avatar Meets Low-Rank Caching, a cascaded framework for real-time one-shot emotion-controllable portrait animation. Our key idea is to decouple controllable motion generation from cross-identity portrait synthesis. We first train a single emotion-aware proxy avatar to generate expressive and controllable driving videos from audio and emotion labels. Although the proxy avatar is trained on one identity, it serves only as a motion source rather than a geometry or appearance template. The subsequent retargeting model, trained on large-scale diverse facial videos, extracts identity-independent motion from the proxy performance and adapts it to arbitrary target portraits. To further improve efficiency, we introduce zero-shot appearance reuse via low-rank caching, which caches reference appearance features at the initial denoising step and models subsequent feature variations with lightweight low-rank adapters. Experiments demonstrate that our method improves emotion controllability while substantially reducing inference cost for real-time portrait animation.

Audio-driven portrait animation has advanced rapidly with diffusion-based generative models, yet real-time one-shot generation with expressive emotion control remains challenging. Existing methods often suffer from insufficient emotion-aware motion priors and expensive appearance computation during multi-step denoising. To address these issues, we propose Proxy Avatar Meets Low-Rank Caching, a cascaded framework for real-time one-shot emotion-controllable portrait animation. Instead of directly generating the target portrait from audio, our method uses a Gaussian-based emotion proxy avatar as a reusable motion generator, which is trained once on a single identity to produce expressive driving videos from audio and emotion labels. Since the proxy avatar only provides motion rather than target appearance or geometry, a large-scale one-shot retargeting model further extracts identity-independent motion from the proxy performance and adapts it to arbitrary target portraits. To improve inference efficiency, we introduce zero-shot appearance reuse with low-rank caching, which caches reference appearance features at the initial denoising step and models subsequent feature variations using lightweight low-rank adapters. Extensive experiments demonstrate that our method achieves stronger emotional expressiveness, better identity-preserving animation, and substantially reduced inference cost, enabling real-time one-shot portrait animation.
\end{abstract}

\begin{figure}[ht!]
  \centering
    \includegraphics[width=\linewidth]{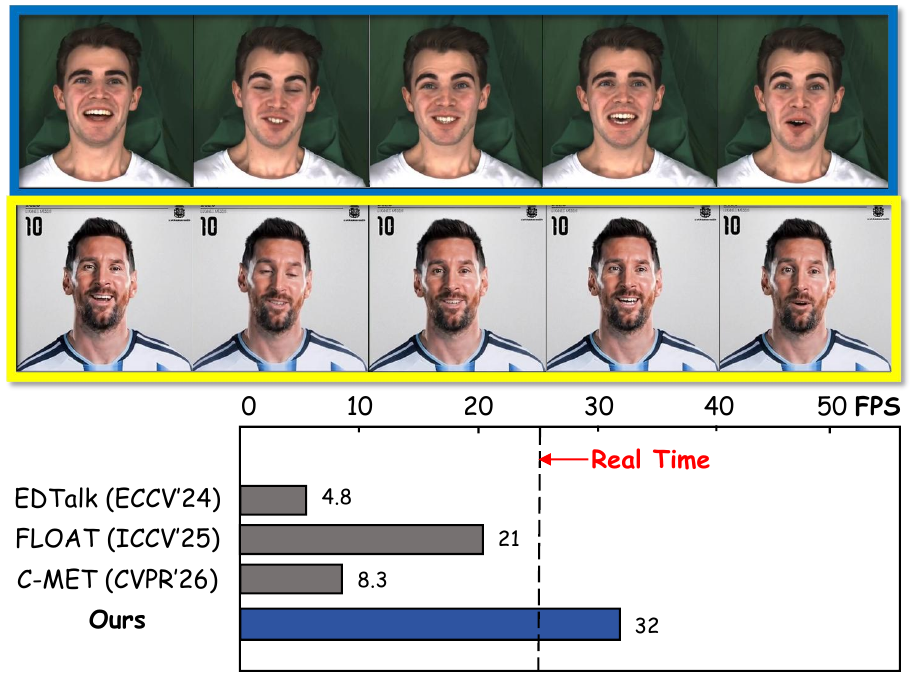}
    \caption{Given a source portrait, an audio clip, and an emotion category, our method can generate emotional talking-head videos in real time on a consumer-grade GPU, specifically the RTX 4090. The blue box indicates the driving videos, and the yellow box indicates the driving results.}
    \label{fig:teaser}
    \vspace{-0.2cm}
\end{figure}

\section{Introduction}

\label{sec:intro}

%交代task和application
Audio-driven talking head generation aims to animate a portrait according to speech, with wide applications in digital humans, and human-computer interaction. 
Fig.~\ref{fig:teaser} shows our target setting of real-time emotion-controllable portrait animation.
Existing methods \cite{wav2lip,vividtalk,hallo} have achieved impressive lip synchronization and visual quality, but rich emotional expressiveness and efficient real-time inference remain challenging.

To achieve emotion control, existing methods have explored different forms of emotional guidance. Early emotion-driven methods \cite{adf,mead} usually inject categorical emotion labels into the generation process, which enables coarse emotion conditioning but often provides limited control over expressive facial dynamics. Some methods \cite{emotalk,consistentavatar} are also person-specific or require training on a limited set of identities, making it difficult to generalize emotional motion to arbitrary portraits. Later works attempt to improve emotional expressiveness by disentangling speech into emotion representations. 
For example, EVP \cite{evp} disentangles speech into content and emotion representations, but unreliable disentanglement may lead to unstable emotion generation from speech alone.
Another line of work uses reference videos to provide emotional or speaking styles, such as EAMM \cite{eamm} and StyleTalk \cite{ma2023styletalk}. Although effective, such guidance is often impractical, since it is difficult to collect reference videos covering diverse emotions and arbitrary speech content. The references may also suffer from mismatched duration, low resolution, or occlusion.
More recently, SPACE \cite{space} and EmoSpeaker \cite{emospeaker} improve controllable expression by introducing explicit emotional conditions and facial representations. Nevertheless, these methods often rely on coefficients, landmarks, or dedicated renderers, and still struggle to jointly achieve flexible emotion control, strong cross-identity generalization, and efficient real-time inference.

To improve inference efficiency, recent talking head systems explore compact motion representations, lightweight renderers, and real-time generation architectures. LivePortrait~\cite{liveportrait} adopts an efficient portrait animation pipeline with stitching and retargeting control. OmniTalker~\cite{omnitalker} further shows that large-scale multimodal training, dual-branch generation, and optimized visual rendering can support real-time audio-video synthesis. More recently, EmoTag~\cite{emotag} introduces a Gaussian-based emotion-aware talking head framework with few-shot personalization, improving rendering efficiency for expressive avatar animation. However, it still relies on identity-specific adaptation and does not address the computational redundancy in diffusion-based one-shot portrait generation. However, diffusion-based portrait animation still repeatedly computes fixed reference appearance features during denoising, making redundant appearance computation a key bottleneck for real-time one-shot generation.

To address these limitations, we revisit the design of emotion-controllable talking head generation from a division-of-labor perspective. Instead of forcing a single model to learn emotional expression, identity adaptation, and efficient synthesis simultaneously, we decompose the problem into three collaborative stages. The first stage builds a Gaussian-based emotion-aware proxy performer, which converts audio and emotion conditions into a vivid intermediate performance with stable articulation and expressive facial dynamics. This intermediate video provides an explicit and controllable motion carrier, but its identity is deliberately not treated as the target appearance. The second stage transfers the proxy performance to an arbitrary portrait through a large-scale one-shot animation model, where the driving signal is interpreted as identity-independent motion rather than proxy-specific geometry. The third stage focuses on the computational bottleneck of diffusion-based synthesis. Since the reference appearance is fixed during generation, we reuse its appearance features across denoising steps and represent the remaining feature changes with low-rank residuals. Although the pipeline appears cascaded, the proxy avatar is trained only once, and each component addresses a distinct bottleneck, forming a coherent and efficient system.

Our contributions are summarized as follows:
\begin{itemize}

    \item We build a Gaussian-based emotion proxy avatar that is trained once per proxy identity and then reused to provide high-quality emotional motion priors.
    \item We use large-scale one-shot retargeting as a bridge to make the single-identity proxy performance reusable for arbitrary target portraits.
    \item We introduce low-rank appearance caching to reuse reference features and approximate denoising variations with low-rank residuals, reducing diffusion inference cost.
\end{itemize}

\begin{figure*}[ht!]
    \centering
    \includegraphics[width=0.9\textwidth]{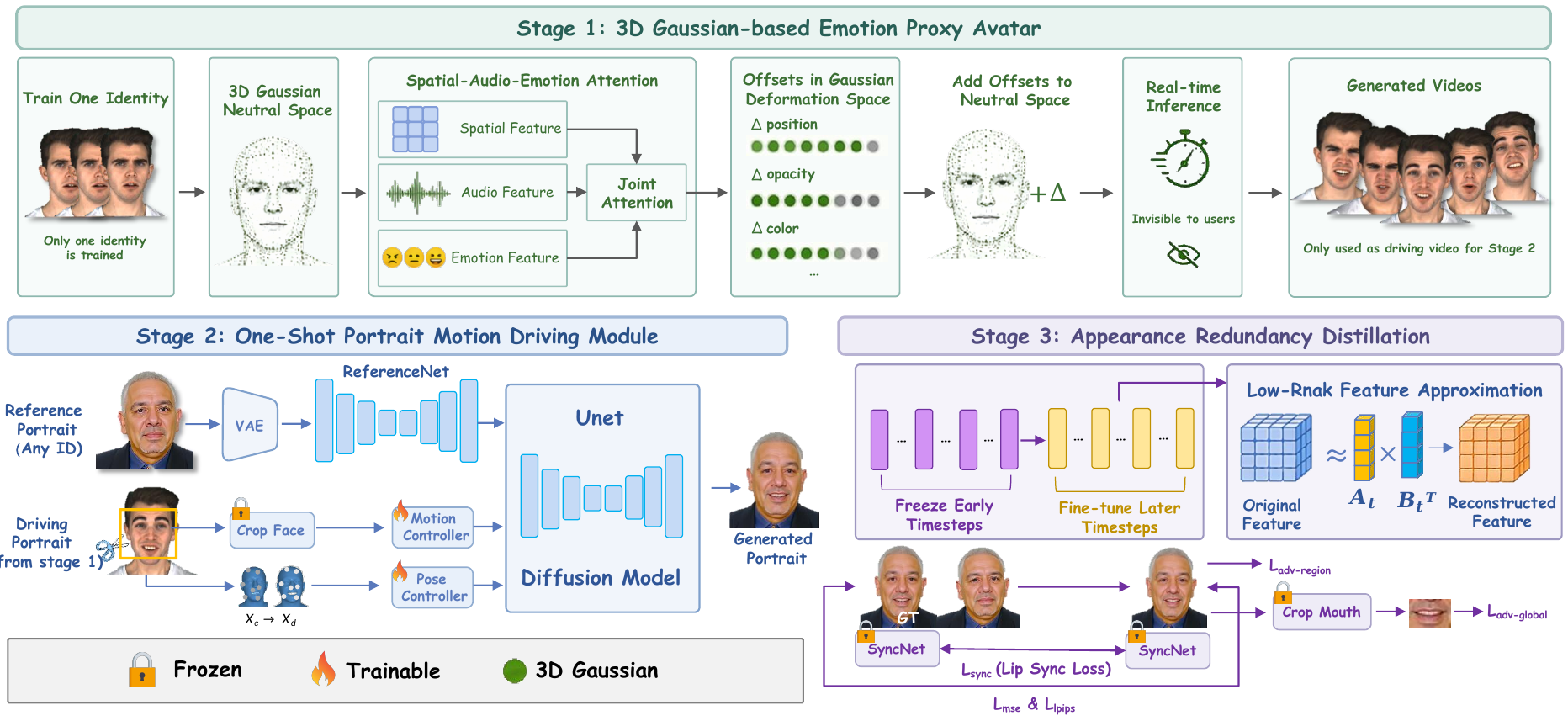}
    \caption{
    \textbf{Overview}. A 3D Gaussian-based emotion proxy avatar first produces controllable emotional driving videos from audio and emotion conditions. 
The one-shot portrait motion driving module then transfers identity-independent proxy motion to arbitrary target portraits. 
Finally, the distillation acceleration module reduces redundant diffusion computation for efficient real-time portrait animation.}
    \label{fig:pipeline}
\end{figure*}

\section{Related Work}
\label{sec:related work}
\subsection{Audio-driven Talking Head Generation}

Audio-driven talking head generation aims to synthesize facial animations from speech and has been widely studied for digital humans and virtual assistants. Early methods mainly focus on lip synchronization, such as Wav2Lip\cite{wav2lip}, while later works improve facial dynamics and head motion through different intermediate representations. For example, MakeItTalk\cite{makelttalk} models speaker-aware motion, Audio2Head\cite{audio2head} generates natural head movements, and SadTalker\cite{sadtalker} predicts 3D motion coefficients for single-image talking face animation. Existing methods can be broadly divided into person-specific and person-independent approaches. Person-specific methods can achieve high fidelity for a given identity but require training or fine-tuning for each subject. In contrast, person-independent methods are trained on large-scale audio-visual datasets and can generalize to unseen portraits. Recent one-shot methods, such as VividTalk\cite{vividtalk} and OmniTalker\cite{omnitalker}, further improve arbitrary-identity animation quality and efficiency. However, controllable emotional expression and real-time inference remain less explored. Our method follows the one-shot setting, but uses an emotion-aware proxy avatar to provide controllable motion priors for arbitrary portrait retargeting.
\subsection{Emotion-controllable Talking Head Generation}
%Emotional expressiveness is crucial for generating natural talking heads. Existing methods usually introduce emotional guidance through labels, speech disentanglement, or reference videos. Label-based methods often rely on discrete emotion categories and emotional datasets such as MEAD \cite{mead}, but their generalization is limited by the scale and diversity of annotated emotional data. EVP\cite{evp} disentangles speech into content and emotion representations for implicit emotion control, but its performance depends on the reliability of such disentanglement. Reference-based methods, such as EAMM\cite{eamm} and StyleTalk\cite{ma2023styletalk}, use emotional or speaking-style videos as guidance. Although intuitive, reference videos are often difficult to obtain in practice, especially when diverse emotions and arbitrary speech content are required. Recent methods further introduce explicit expression conditions or intermediate representations. For example, SPACE\cite{space} enables controllable expression, EmoSpeaker\cite{emospeaker} predicts emotion-related 3D facial coefficients, and EMOdiffhead\cite{emodiffhead} uses FLAME expression vectors to guide diffusion-based synthesis. However, these methods still rely on emotion-specific data, reference videos, coefficient-based representations, or expensive generative models. In contrast, our method learns emotional motion through a single Gaussian-based proxy avatar and transfers identity-independent motion to arbitrary portraits without external emotional reference videos or multi-identity emotion-avatar training.

Emotional expressiveness is essential for natural talking-head generation. Existing methods introduce emotion guidance through labels, reference videos, or explicit facial priors. Label-based methods built on emotional datasets such as MEAD~\cite{mead} provide controllable categories but are limited by annotated data diversity. EVP~\cite{evp} disentangles speech into content and emotion representations, while reference-based methods such as EAMM~\cite{eamm} and StyleTalk~\cite{ma2023styletalk} transfer emotional styles from external videos, which are often difficult to obtain for arbitrary speech and emotions. Recent methods further use explicit intermediate representations: EmoSpeaker~\cite{emospeaker} predicts emotion-related 3D facial coefficients, EMOdiffhead~\cite{emodiffhead} adopts FLAME expression vectors for diffusion-based synthesis, and C-MET~\cite{cmet} learns cross-modal emotion transfer between speech and visual feature spaces for emotion editing. Despite their progress, these methods still depend on emotion-specific data, reference sources, coefficient-based priors, or costly generative models. In contrast, our Gaussian-based proxy avatar provides reusable identity-independent emotional motion without external emotional videos or multi-identity avatar training.

\subsection{Efficient Portrait Animation}

Efficiency is essential for real-time talking head generation. Recent methods improve speed with compact motion representations, lightweight renderers, or optimized generation pipelines. For example, LivePortrait\cite{liveportrait} achieves efficient portrait animation with stitching and retargeting control, VASA-1\cite{vasa-1} generates lifelike talking faces in real time, and OmniTalker\cite{omnitalker} explores real-time multimodal talking head generation with an optimized visual renderer. However, these methods mainly rely on lightweight rendering pipelines rather than diffusion-based high-fidelity synthesis. Diffusion-based methods, such as EchoMimic\cite{echomimic}, EMO\cite{emo}, and EMOdiffhead\cite{emodiffhead}, improve realism and expressiveness by progressively denoising visual latents under audio, identity, or expression conditions. Nevertheless, iterative denoising introduces substantial computational cost. In portrait animation, the reference image remains fixed, and its appearance features are highly reusable across frames and denoising steps. Existing diffusion-based methods rarely exploit this redundancy. So we reuse reference appearance features with low-rank caching, reducing inference cost while preserving visual fidelity.

\section{Method}
\label{sec:method}

%Given a reference portrait, a driving audio clip, and an emotion condition, our goal is to generate a realistic talking-head video with accurate lip synchronization, controllable emotional expression, and real-time inference. As shown in Fig.~\ref{fig:pipeline}, our framework consists of three stages. The following sections describe these three components in order: the emotion proxy avatar, the one-shot portrait motion driving module, and the appearance redundancy distillation strategy.

Given a reference portrait, a driving audio clip, and an emotion condition, our goal is to generate a talking-head video with accurate lip synchronization, controllable emotional expression, and real-time inference. As shown in Fig.~\ref{fig:pipeline}, our framework comprises three stages, which are introduced in the following order: emotion proxy avatar, one-shot portrait motion driving, and appearance redundancy distillation.

\subsection{3D Gaussian-based Emotion Proxy Avatar}
\label{sec:proxy_avatar}

The first stage aims to build an emotion-controllable proxy performer that converts audio and emotion conditions into an explicit driving video. Instead of learning emotional motion for arbitrary identities directly, we train the proxy avatar once on a single identity. This design concentrates audio-emotion motion learning in a controllable proxy space, while leaving cross-identity adaptation to the following one-shot portrait motion driving module.

\noindent\textbf{Gaussian Proxy Representation.}
We represent the proxy avatar with a set of 3D Gaussians:
\begin{equation}
    \mathcal{G} = \{ \mathbf{x}_i, \mathbf{r}_i, \mathbf{s}_i, \mathbf{c}_i, \alpha_i \}_{i=1}^{N},
\end{equation}
where $\mathbf{x}_i$, $\mathbf{r}_i$, $\mathbf{s}_i$, $\mathbf{c}_i$, and $\alpha_i$ denote the center position, rotation, scale, color, and opacity of the $i$-th Gaussian, respectively. Each Gaussian is associated with a covariance matrix:
\begin{equation}
    \mathbf{\Sigma}_i = \mathbf{R}_i \mathbf{S}_i \mathbf{S}_i^{\top} \mathbf{R}_i^{\top},
\end{equation}
where $\mathbf{R}_i$ and $\mathbf{S}_i$ are derived from $\mathbf{r}_i$ and $\mathbf{s}_i$. Given a camera view, the Gaussians are projected onto the image plane and rendered by alpha compositing:
\begin{equation}
    \mathbf{C}(\mathbf{p}) =
    \sum_{i \in \mathcal{N}(\mathbf{p})}
    T_i \alpha_i \mathbf{c}_i,
    \quad
    T_i = \prod_{j<i}(1-\alpha_j),
\end{equation}
where $\mathbf{p}$ is a pixel location, $\mathcal{N}(\mathbf{p})$ denotes the depth-ordered Gaussians contributing to this pixel, and $T_i$ is the accumulated transmittance. This explicit representation enables high-fidelity and real-time rendering, making it suitable for constructing the proxy performer.

\noindent\textbf{Audio-Emotion Motion Decomposition.}
Given an audio clip $\mathbf{A}=\{a_t\}_{t=1}^{T}$ and an emotion condition $\mathbf{e}$, where $t$ is the frame-aligned time step, the proxy avatar predicts frame-wise facial motion. We first encode audio and emotion as:
\begin{equation}
    \mathbf{f}^{a}_t = \mathcal E_a(a_t),
    \quad
    \mathbf{f}^{e} = \mathcal E_e(\mathbf{e}),
\end{equation}
where $\mathcal E_a(\cdot)$ is the audio encoder and $\mathcal E_e(\cdot)$ is the emotion embedding layer. The audio feature mainly provides phoneme-related articulation cues, while the emotion feature controls expression-related facial dynamics. To stabilize emotional motion generation, we decompose the proxy motion into neutral speech motion and emotion-dependent residual motion:
\begin{equation}
    \mathbf{m}_t = \mathbf{m}^{neu}_t + \Delta \mathbf{m}^{emo}_t.
\end{equation}
Here, $\mathbf{m}^{neu}_t$ captures audio-synchronized talking motion, and $\Delta \mathbf{m}^{emo}_t$ models affective facial deformation. This decomposition prevents the emotion branch from relearning phoneme-driven mouth motion and encourages it to focus on brows, cheeks, eyelids, and mouth-corner dynamics.

\noindent\textbf{Spatial-Audio-Emotion Joint Attention.}
To capture the mutual dependency among facial regions, speech content, and emotional states, we introduce a spatial-audio-emotion joint attention module. The spatial-audio-emotion attention module consists of multiple sets of cross-attention layers $F_{ca}$ and feed-forward layers $F_{fd}$, each connected via skip connections. The module is formulated as follows:
\begin{align}
&z_{n}^{0}=q_n,\\
&z_{n}^{\prime l}={F}_{ca}(z_{n}^{l-1},\mathbf{f}^{a}_t,\mathbf{f}^{e})+z_{n}^{l-1},\quad l=1...L,\\
&z_{n}^{l}=F_{fd}(z_{n}^{\prime l})+z_{n}^{\prime l},\quad l=1...L.
\end{align}
 By computing the cross-attention between the spatial query $q_n$, audio features $\mathbf{f}^{a}_t$ and emotion features $\mathbf{f}^{e}$ of the $t^{th}$ image frame, the output features successfully integrate these signals with the rich facial details captured by the Gaussian model, preparing for subsequent Gaussian deformation in the emotion space. 
 
 \noindent\textbf{Emotion-aware Gaussian Deformation.}
 To predict the offsets for each Gaussian attribute, we use a set of MLP regressors, $\mathcal{F}_d$, as detailed below:
\begin{align}
(\Delta \mathbf{x}_{i,t},\Delta \mathbf{r}_{i,t},\Delta \mathbf{s}_{i,t},\Delta \mathbf{c}_{i,t},\Delta \alpha_{i,t})=\mathcal{F}_d(z_{n}^{L}).
\end{align}
Therefore, the Gaussian attributes in the final emotional space are: $\mathbf{x}^{emo}_{i,t} = \mathbf{x}^{neu}_{i,t} + \Delta \mathbf{x}_{i,t}$,
    $\mathbf{r}^{emo}_{i,t} = \mathbf{r}^{neu}_{i,t} + \Delta \mathbf{r}_{i,t}$, 
    $\mathbf{s}^{emo}_{i,t} = \mathbf{s}^{neu}_{i,t} + \Delta \mathbf{s}_{i,t}$, 
    $\mathbf{c}^{emo}_{i,t} = \mathbf{c}^{neu}_{i,t} + \Delta \mathbf{c}_{i,t}$,
    $\alpha^{emo}_{i,t} = \alpha^{neu}_{i,t} + \Delta \alpha_{i,t}$.
Training objectives are provided in the \textbf{appendix}. After training, the proxy avatar renders an emotional driving video in real time:
\begin{equation}
    \mathbf{V}^{p}
    =
    \mathcal{R}_{G}
    (\mathbf{A}, \mathbf{e}; \Theta_{p}),
\end{equation}
where $\mathcal{R}_{G}$ denotes the Gaussian rendering process and $\Theta_{p}$ represents the learned proxy avatar parameters. The proxy video $\mathbf{V}^{p}$ contains synchronized lip motion and emotion-aware facial dynamics, it serves as an explicit motion carrier for the next stage, where identity-independent motion is extracted and retargeted to an arbitrary reference portrait.

\subsection{One-Shot Portrait Motion Driving Module}
\label{sec:one_shot_driving}

%Given the proxy driving video generated by the 3D Gaussian-based emotion proxy avatar, the second stage aims to transfer its emotional facial motion to an arbitrary reference portrait $I_R$. 
%We denote the proxy driving video as $\mathbf{V}^{p}=\{I^{p}_{t}\}_{t=1}^{T}$ and the frame-aligned audio features as $\mathbf{A}=\{a_t\}_{t=1}^{T}$, where $I^{p}_{t}$ and $a_t$ correspond to the same timestamp. Although the proxy video provides expressive lip motion and emotion-aware facial dynamics, it still carries the geometry and appearance of the proxy identity. Therefore, we do not directly copy its pixels or facial shape. Instead, we extract identity-independent motion cues from the proxy video and inject them into a one-shot portrait animation model conditioned on the target reference image.

Given the proxy driving video generated by the 3D Gaussian-based emotion proxy avatar, the second stage transfers its emotional facial motion to an arbitrary reference portrait $I_R$. We denote the proxy driving video as $\mathbf{V}^{p}=\{I^{p}_{t}\}_{t=1}^{T}$ and the frame-aligned audio features as $\mathbf{A}=\{a_t\}_{t=1}^{T}$, where $I^{p}_{t}$ and $a_t$ correspond to the same timestamp. Although the proxy video contains expressive lip motion and emotion-aware facial dynamics, it still reflects the proxy identity's geometry and appearance. Therefore, we extract identity-independent motion cues from the proxy video and inject them into a one-shot portrait animation conditioned on $I_R$.

\noindent\textbf{Controller-Guided Portrait Generation.}
We adopt a diffusion-based portrait animation backbone to synthesize the target frame $I_t$ from a noisy latent $\mathbf{z}_t^{\tau}$ at denoising step $\tau$. The generation process is conditioned on the reference appearance, proxy motion, proxy pose:
\begin{equation}
    \hat{\epsilon}
    =
    \epsilon_{\theta}
    \left(
    \mathbf{z}_t^{\tau}, \tau;
    \mathbf{c}^{app},
    \mathcal{C}_t
    \right),
    \quad
    \mathcal{C}_t =
    \{\mathbf{c}^{mot}_t,\mathbf{c}^{pose}_t\},
\end{equation}
where $\mathbf{c}^{app}$ is extracted from the reference image $I_R$, $\mathbf{c}^{mot}_t$ and $\mathbf{c}^{pose}_t$ are obtained from the proxy frame $I^p_t$. The reference appearance condition is computed by a ReferenceNet:
\begin{equation}
    \mathbf{c}^{app} = \mathcal{R}(I_R),
\end{equation}
which provides identity, texture, and background information for preserving the target portrait.

\noindent\textbf{Motion Controller.}
The motion controller captures fine-grained facial dynamics from the proxy video, including mouth shapes, eye movements, and local expression changes. Specifically, we use a motion encoder $\mathcal{E}_{m}$ to extract implicit facial motion representations:
\begin{equation}
    \mathbf{c}^{mot}_t = \mathcal{E}_{m}(I^{p}_{t}).
\end{equation}
These implicit features are injected into the spatial layers of the denoising network through cross-attention:
\begin{equation}
    \mathrm{Attn}_{mot}
    =
    \mathrm{Softmax}
    \left(
    \frac{\mathbf{Q}_{z}\mathbf{K}_{m}^{\top}}{\sqrt{d}}
    \right)
    \mathbf{V}_{m},
\end{equation}
where $\mathbf{Q}_{z}$ is projected from the denoising feature, while $\mathbf{K}_{m}$ and $\mathbf{V}_{m}$ are projected from $\mathbf{c}^{mot}_t$. This controller transfers local facial dynamics from the proxy performance while avoiding direct dependence on the proxy appearance.

\noindent\textbf{Pose Controller.}
The motion controller mainly focuses on local expression dynamics, but global head pose, translation, and scale also need to be aligned with the target portrait. Therefore, we introduce a pose controller based on 3D implicit keypoints. Following the one-shot retargeting formulation, we extract 3D motion parameters from both the proxy frame and the reference portrait:
\begin{equation}
    \mathbf{k}^{p}_{c,t}, \mathbf{R}^{p}_{t}, \mathbf{t}^{p}_{t}, s^{p}_{t}
    =
    \mathcal{E}_{k}(I^{p}_{t}),
\end{equation}
\begin{equation}
    \mathbf{k}^{R}_{c}, \mathbf{R}^{R}, \mathbf{t}^{R}, s^{R}
    =
    \mathcal{E}_{k}(I_R),
\end{equation}
where $\mathbf{k}^{R}_{c}$ denotes the canonical 3D keypoints of the reference portrait, encoding the target facial geometry such as face shape and facial structure. $\mathbf{R}^{R}$, $\mathbf{t}^{R}$, and $s^{R}$ represent the head rotation, translation, and scale of the reference portrait, respectively.
To adapt the proxy pose to the target geometry, we compute the target-space driving keypoints:
\begin{equation}
    \mathbf{k}^{d}_{t}
    =
    s^{p}_{t} \cdot \mathbf{k}^{R}_{c} \mathbf{R}^{p}_{t}
    +
    \mathbf{t}^{p}_{t}.
\end{equation}
The resulting keypoints are mapped into a spatial condition map by a pose guider:
\begin{equation}
    \mathbf{c}^{pose}_t = \mathcal{P}(\mathbf{k}^{d}_{t}),
\end{equation}
and injected into the denoising backbone through spatial addition. This controller provides explicit geometric guidance for head movement and coarse facial layout, complementing the implicit motion features.

After training on large-scale diverse facial videos, the one-shot portrait motion driving module learns to extract identity-independent motion from the proxy performance and adapt it to arbitrary target portraits. Training
objectives are provided in the \textbf{appendix}. The generated portrait sequence from this stage serves as the high-quality diffusion synthesis result, while the next stage further reduces its inference cost through appearance-redundancy-based distillation acceleration.

\subsection{Appearance Redundancy Distillation}
\label{sec:appearance_distillation}

Although the one-shot portrait motion driving module can generate high-fidelity animated portraits, its diffusion-based generation process still requires multiple denoising steps and repeated appearance conditioning. In one-shot portrait animation, the target identity is specified by a fixed reference image $I_R$. Therefore, its identity, texture, and background appearance remain largely unchanged throughout generation. However, conventional diffusion-based portrait animation models repeatedly extract and inject reference appearance features at different denoising steps, resulting in substantial redundant computation. To address this issue, we propose an appearance redundancy distillation strategy that compresses the denoising process and reuses reference appearance features with lightweight low-rank residual adaptation.

\noindent\textbf{Few-step Appearance Distillation.}
Based on the observation reported in the \textbf{appendix} that portrait structure and motion are mainly established in the early denoising steps, we first distill the original multi-step sampling process into a compact schedule:
\begin{equation}
    \mathcal{T}_{K} = \{\tau_1,\tau_2,\dots,\tau_K\},
    \quad K \ll N,
\end{equation}
where $N$ is the number of denoising steps in the original diffusion model and $K$ is the number of distilled steps. Starting from Gaussian noise $\mathbf{z}^{\tau_K}$, the student model progressively denoises the latent along $\mathcal{T}_{K}$ and predicts the clean latent $\hat{\mathbf{z}}^0$, which is decoded into the image space:
\begin{equation}
    \hat{I}_t = \mathcal{D}_{vae}(\hat{\mathbf{z}}^0_t),
\end{equation}
where $\mathcal{D}_{vae}$ is the VAE decoder. The distilled model is supervised by a hybrid objective $\mathcal{L}_{step}$, see \textbf{appendix}.

\begin{comment}
\begin{equation}
    \mathcal{L}_{step}
    =
    \|\hat{I}_t-I_t\|_2^2
    +
    \lambda_{lpips}\mathcal{L}_{lpips}(\hat{I}_t,I_t)
    +
    \lambda_{adv}^{g}\mathcal{L}_{adv}^{g}(\hat{I}_t)  
    +
    \lambda_{sync}\mathcal{L}_{sync}(\hat{\mathbf{V}},\mathbf{A})
    +
    \lambda_{adv}^{m}\mathcal{L}_{adv}^{m}(\mathcal{M}(\hat{I}_t)),
\end{equation}
where $\hat{I}_t$ denotes the generated frame at time step $t$, and $I_t$ denotes the corresponding ground-truth frame. The term $\|\hat{I}_t-I_t\|_2^2$ is the pixel-level reconstruction loss. $\mathcal{L}_{lpips}$ denotes the perceptual loss for preserving visual fidelity. $\mathcal{L}_{adv}^{g}$ is the global adversarial loss applied to the whole generated frame. $\hat{\mathbf{V}}$ denotes the generated video sequence, and $\mathbf{A}$ denotes the corresponding audio condition. $\mathcal{L}_{sync}$ is the SyncNet-based lip synchronization loss, which encourages the generated mouth motion to be aligned with the input speech. $\mathcal{M}(\cdot)$ denotes the mouth-region cropping operation, and $\mathcal{L}_{adv}^{m}$ is the local mouth adversarial loss for improving fine-grained mouth realism. $\lambda_{lpips}$, $\lambda_{adv}^{g}$, $\lambda_{sync}$, and $\lambda_{adv}^{m}$ are the balancing weights for the corresponding loss terms.
\end{comment}

\noindent\textbf{Low-rank Appearance Reuse.}
Step distillation reduces the number of denoising steps, but each step may still require costly reference appearance extraction. To further remove appearance redundancy, we compute the full reference appearance feature only at the first distilled step:
\begin{equation}
    \mathbf{F}_1 = \mathcal{R}_{\tau_1}(I_R).
\end{equation}
For the following denoising steps, instead of re-running the full appearance extractor, we approximate the reference feature as:
\begin{equation}
    \widetilde{\mathbf{F}}_j
    =
    \mathbf{F}_1 + \Delta \mathbf{F}_j,
    \quad i=2,\dots,K.
\end{equation}
Since the difference between appearance features across denoising steps is small and structured, we parameterize the residual $\Delta \mathbf{F}_j$ with a low-rank decomposition:
\begin{equation}
    \Delta \mathbf{F}_j
    \approx
    \mathbf{A}_j \mathbf{B}_j,
\end{equation}
where $\mathbf{A}_j \in \mathbb{R}^{d \times r}$ and $\mathbf{B}_j \in \mathbb{R}^{r \times m}$ are lightweight learnable matrices, with $r \ll \min(d,m)$. For a multi-layer appearance feature pyramid, the approximation is applied at each layer:
\begin{equation}
    \widetilde{\mathbf{F}}^{l}_j
    =
    \mathbf{F}^{l}_1
    +
    \mathbf{A}^{l}_j \mathbf{B}^{l}_j,
    \quad l=1,\dots,L.
\end{equation}
The denoising network uses $\widetilde{\mathbf{F}}_j$ as the appearance condition:
\begin{equation}
    \hat{\epsilon}_j
    =
    \epsilon_{\theta}
    \left(
    \mathbf{z}_t^{\tau_j}, \tau_j;
    \mathcal{C}_t,
    \widetilde{\mathbf{F}}_j
    \right).
\end{equation}
This design preserves step-specific appearance adaptation while avoiding repeated full ReferenceNet computation.

\noindent\textbf{Appearance Feature Distillation.}
To ensure that the low-rank reused feature approximates the original full appearance feature, we introduce an appearance feature distillation loss. Let $\mathbf{F}^{l}_j$ be the full reference feature extracted by the original appearance branch at step $\tau_j$, and let $\widetilde{\mathbf{F}}^{l}_j$ be the low-rank approximated feature. We minimize:
\begin{equation}
    \mathcal{L}_{app}
    =
    \sum_{j=2}^{K}
    \sum_{l=1}^{L}
    \left\|
    \widetilde{\mathbf{F}}^{l}_j
    -
    \mathrm{sg}(\mathbf{F}^{l}_j)
    \right\|_1,
\end{equation}
where $\mathrm{sg}(\cdot)$ denotes the stop-gradient operation. This loss transfers the behavior of the full appearance extractor to the low-rank residual branch without updating the teacher appearance features.
The final distillation objective is:
\begin{equation}
    \mathcal{L}_{ARD}
    =
    \mathcal{L}_{step}
    +
    \lambda_{app}\mathcal{L}_{app}.
\end{equation}

\section{Experiments}
\noindent\textbf{Implementation Details.}
For the 3D Gaussian-based emotion proxy avatar, we select one subject from MEAD~\cite{mead} with multiple emotion categories and all three intensity levels, and train a single proxy avatar once. Instead of training separate proxy avatars for different intensity levels, we use all intensity levels jointly to enrich emotional motion diversity, which reduces training overhead and avoids maintaining multiple intensity-specific models. During inference, the proxy motion is conditioned on the input audio and emotion category, where the audio dynamics naturally modulate the strength of the generated emotional motion. The proxy avatar is implemented in PyTorch and trained with the Adam optimizer \cite{adam}
, where the learning rate is initialized as $1\times10^{-4}$ and gradually decayed to $1\times10^{-5}$. For the one-shot portrait motion driving and appearance redundancy distillation modules, we follow PersonaLive~\cite{personalive} and train on VFHQ~\cite{vfhq}, NerSemble~\cite{nersemble}, and DH-FaceVid-1K~\cite{dh}. All videos are processed at 25 fps and cropped to $512 \times 512$. For adversarial training, we adopt the StyleGAN2 discriminator initialized with FFHQ-pretrained weights~\cite{ffhq}. The denoising steps are set to $N=4$, the chunk size is set to $M=4$, and the motion threshold in Historical Keyframe Mechanism (HKM) is set to $\tau=17$. The model is trained on two A100 GPUs using AdamW \cite{adamw} with a learning rate of $1\times10^{-5}$ and a weight decay of $0.01$. For self-reconstruction, we train one proxy avatar for each of two selected MEAD identities, and each proxy is trained only once using all emotion categories and intensity levels. This setting is used solely as an identity-specific diagnostic to verify reconstruction robustness across subjects; it does not require retraining per test clip or per emotion and does not change our main one-shot driving pipeline. For cross-identity evaluation, we randomly select 10 target identities and report the averaged results over all generated videos.

\noindent\textbf{Evaluation Metrics.}
We adopt different metrics for self-reconstruction and cross-identity driving. For self-reconstruction, we report PSNR~\cite{PSNR}, LPIPS~\cite{LPIPS}, SSIM~\cite{SSIM}, Landmark Distance (LMD), Action Unit Error (AUE) for lower/upper facial regions, SyncNet-based Lip Sync~\cite{SyncNet}, and FPS. For cross-identity driving, where paired ground truth is unavailable, we report FID~\cite{FID}, identity similarity (ID-Sim), emotion classification accuracy (Emo-Acc), and Lip Sync \cite{SyncNet}.

%MimicTalk nips24 开源
%TalkingGaussian eccv 24 开源
%InsTaG cvpr 25 开源
%EmoTaG cvpr2026 开源
\noindent\textbf{Comparison Settings.}
We evaluate our method under two settings. 
(1) \textbf{Self-reconstruction}, where a reference image and the corresponding audio from the same identity are used to reconstruct the target video.   
(2) \textbf{One-shot cross-identity driving}, where the target portrait is provided by a single reference image, while the driving audio and emotion condition come from a different identity.

\begin{figure}[ht!]
  \centering
    \includegraphics[width=0.9\linewidth]{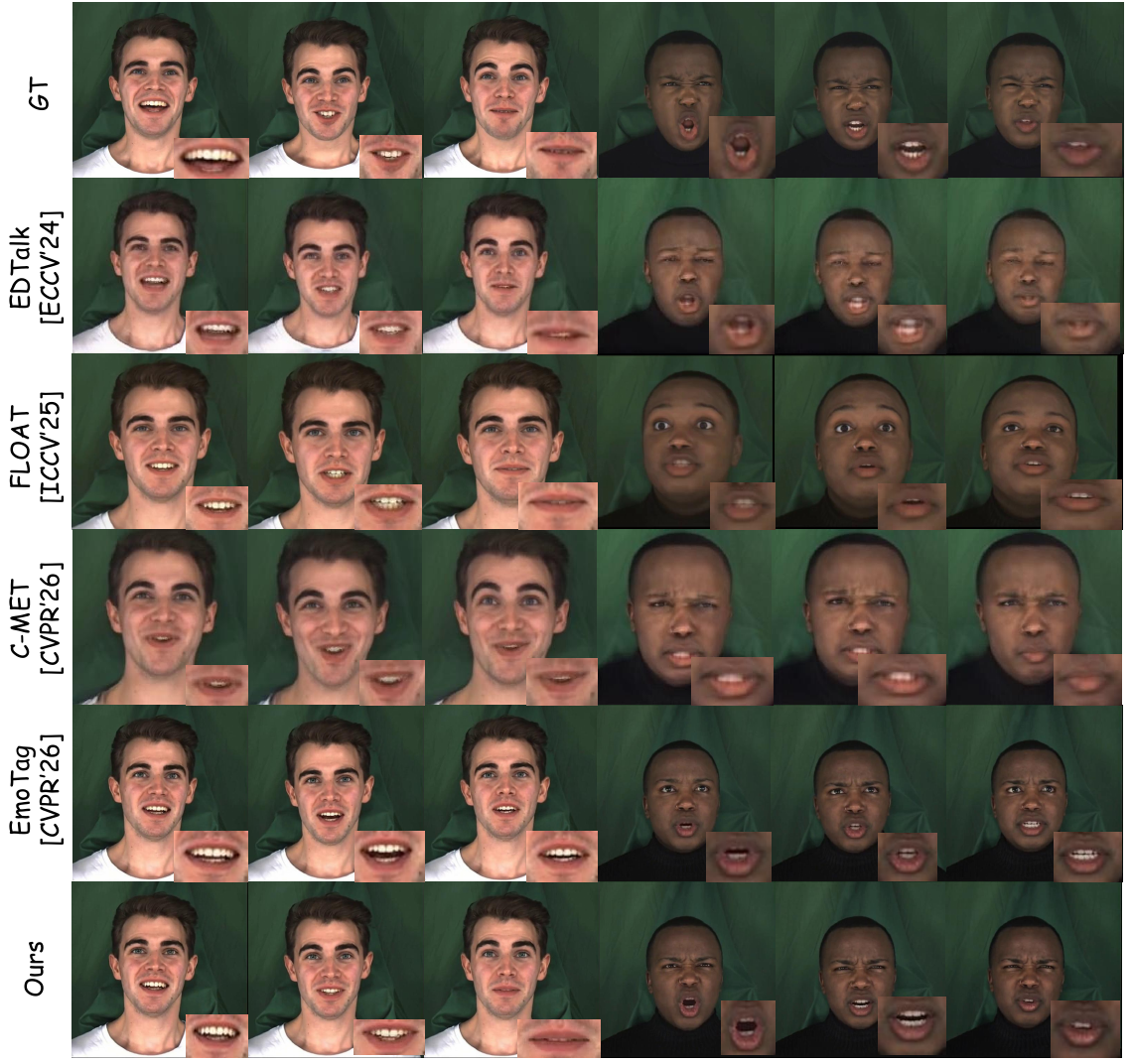}
    \caption{
Qualitative comparison on \textbf{self-reconstruction}. Our method generates more expressive, visually faithful, and well-synchronized talking heads than previous methods across emotional cases.}
    \label{fig:self_recon}
\end{figure}

\begin{figure}[ht!]
  \centering
    \includegraphics[width=0.9\linewidth]{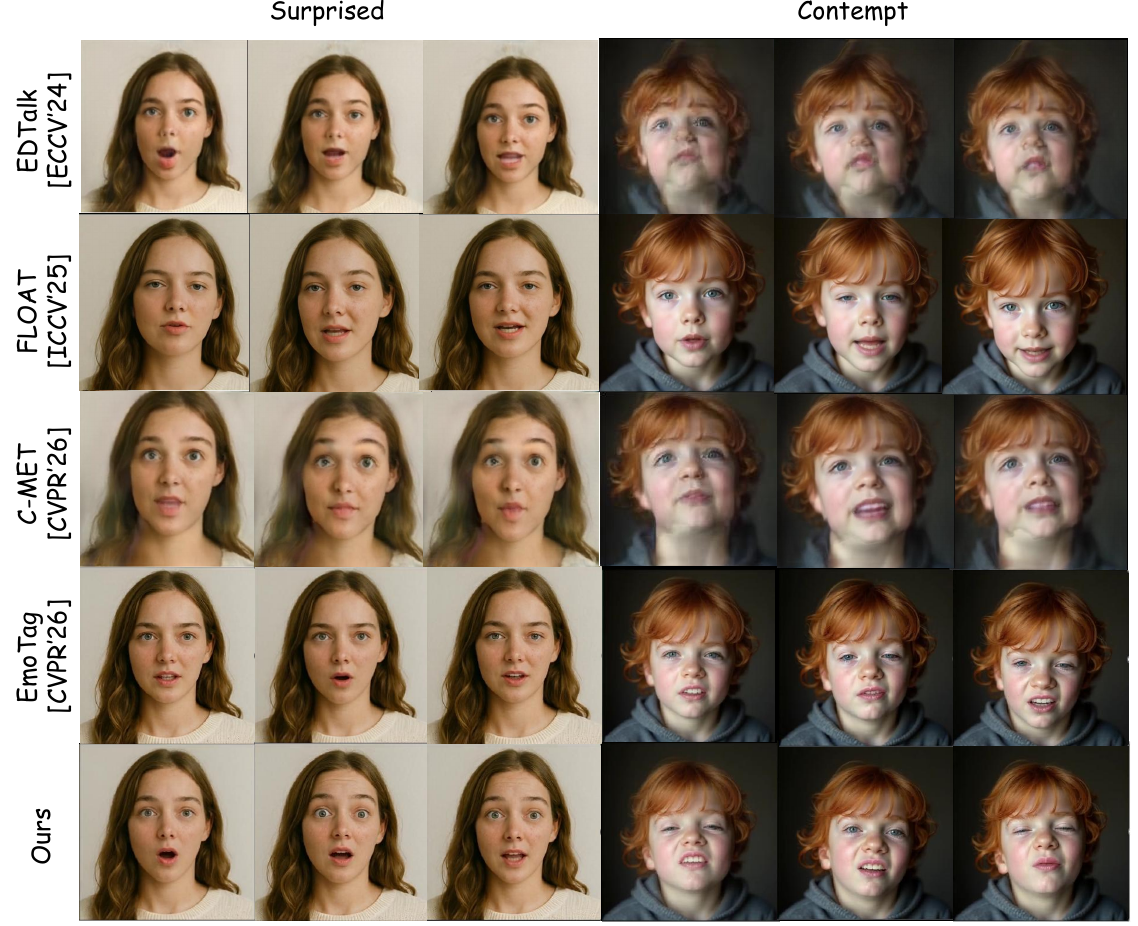}
    \caption{
Qualitative comparison on one-shot \textbf{cross-identity} driving. Given a single reference portrait and driving audio from a different identity, our method produces more expressive emotional facial motions while better preserving the target identity. 
}
    \label{fig:cross_identity}
\end{figure}

\noindent\textbf{Qualitative Comparison.}
Fig. \ref{fig:self_recon} shows the qualitative comparison under the self-reconstruction setting. Compared with existing methods, our method generates more expressive results while maintaining stable audio-lip synchronization. EDTalk \cite{edtalk} and C-MET \cite{cmet} suffer from identity inconsistency and audio mismatch in some cases, while FLOAT\cite{float} lacks natural emotional variations. Although EmoTaG\cite{emotag} produces reasonable talking-head animations, its emotional expressiveness is still limited, especially when generating high-intensity emotional expressions. Fig. \ref{fig:cross_identity} shows the qualitative comparison under the one-shot cross-identity driving setting. Compared with existing methods, our method better transfers emotion-aware facial motion to unseen target portraits while preserving the target identity. The generated results exhibit more vivid emotional expressions, more natural facial dynamics, and fewer identity distortions, demonstrating the advantage of our identity-independent motion extraction and one-shot retargeting design. Please refer to the \textbf{demo} for dynamic comparisons.

\noindent\textbf{Long-video Stability.}
We further evaluate long-video generation stability in the supplementary \textbf{demo}. Our method maintains stable identity, coherent facial motion, and consistent lip synchronization over extended sequences, without obvious temporal drift.

\begin{table}[t!]\small
    \centering
   \setlength{\tabcolsep}{1mm}

    \resizebox{0.9\linewidth}{!}{
    \begin{tabular}{c|cccccccc}
        \midrule
        Method&PSNR$\uparrow$&LPIPS$\downarrow$&SSIM$\uparrow$&LMD$\downarrow$&AUE-(L/U)$\downarrow$&Lip Sync$\uparrow$&FPS$\uparrow$\\
        \midrule
        EDTalk-ECCV'24 & 27.38&0.32& 0.745& 3.16 & 0.841-0.347 & 5.655 & 4.8\\
        FLOAT-ICCV'25  &26.46&0.43 &0. 712 & 3.67 & 0.992-0.691 &4.913 & 21 \\
        C-MET-CVPR'26 &26.10& 0.49 & 0.704 & 4.29 & 1.135-0.829&3.540 & 8.3 \\
        EmoTag-CVPR'26 &28.31 & 0.021 & 0.764 & 2.83 & 0.695-0.277 & 5.982 & \textbf{45} \\
        w/o PA & 29.02 & 0.030 & 0.832 & 2.76 & 0.811-0.318 & 5.783 & 32 \\
        w/o LRC + w/o few-step & \textbf{31.42} & \textbf{0.017} & \textbf{0.890} & \textbf{2.23} & \textbf{0.664-0.198} & \textbf{6.351} & 7.8 \\
        w/o LRC + w/ few-step & 31.28 & 0.018 & 0.887 & 2.26 & 0.670-0.204 & 6.322 & 21.5 \\
        
        \textbf{Ours} & 31.24 & 0.018 & 0.883 & 2.27 & 0.672-0.207 & 6.314 & 32 \\

        \bottomrule

    \end{tabular}}
               \caption{
        Quantitative comparison and ablation study under the self-reconstruction setting.
        }
    \label{tab:self-recon-tab}
    
\end{table}

\noindent\textbf{Quantitative Results.}
Tab. \ref{tab:self-recon-tab} reports the quantitative comparison under the self-reconstruction setting. Our method achieves the best performance on most metrics (excluding ablation study), demonstrating superior visual fidelity, more accurate facial motion, and better audio-lip synchronization. Although EmoTag \cite{emotag} achieves the highest FPS, our method still runs at 32 FPS, satisfying real-time inference while providing significantly better generation quality and motion accuracy. Tab. \ref{tab:cross_identity_quantitative} reports the quantitative results under the one-shot cross-identity driving setting. Since paired ground-truth videos are unavailable in this setting, we evaluate distribution-level visual quality, identity preservation, emotional expressiveness, and lip synchronization. Our method achieves the best performance across all metrics, with the lowest FID and the highest ID-Sim, Emo-Acc, and Lip Sync scores. These results demonstrate that our method can better preserve the target identity while transferring expressive emotion-aware facial motion and maintaining accurate audio-lip synchronization.

\begin{table}[t]
    \centering

    \resizebox{0.9\linewidth}{!}{
    \begin{tabular}{ccccc}
        \toprule
        Method & FID$\downarrow$ & ID-Sim$\uparrow$ & Emo-Acc$\uparrow$ & Lip Sync$\uparrow$ \\
        \midrule
        EDTalk-ECCV'24   & 42.37 & 0.612 & 55.8 & 5.214 \\
        FLOAT-ICCV'25    & 47.85 & 0.583 & 48.7 & 4.766 \\
        C-MET-CVPR'26    & 50.92 & 0.561 & 51.3 & 3.182 \\
        EmoTag-CVPR'26   & 34.91 & 0.702 & 63.2 & 5.742 \\
        w/o PA & 35.87 & 0.721 & 61.4 & 5.632 \\
        w/o LRC + w/o few-step & \textbf{29.41} & \textbf{0.752} & \textbf{76.2} & \textbf{6.141} \\
        w/o LRC + w/ few-step & 29.57 & 0.749 & 75.6 & 6.105 \\

        \textbf{Ours}    & 29.64 & 0.745 & 75.4 & 6.096 \\
        \bottomrule
    \end{tabular}
    }
        \caption{
    Quantitative comparison and ablation study under the one-shot cross-identity driving. 
    }
    \label{tab:cross_identity_quantitative}
\end{table}

\noindent\textbf{User Study.} 
To further evaluate perceptual quality, we conduct a user study against representative baseline methods. We invite 20 participants to assess anonymized videos from the self-reconstruction setting in terms of emotional expressiveness, lip synchronization, and visual realism. As reported in Tab \ref{tab:user_study}, our method obtains the highest overall ratings, especially in emotional expressiveness, indicating that it produces more vivid and coherent emotion-aware facial motions.

\begin{table}[t]
    \centering

    \resizebox{0.9\linewidth}{!}{
    \begin{tabular}{cccc}
        \toprule
        Method & Emo Exp$\uparrow$ & Lip Syn$\uparrow$ & Visual Realism$\uparrow$\\
        \midrule
        EDTalk-ECCV'24  & 5\% &0\% & 15\% \\
        FLOAT-ICCV'25   & 10\%& 5\% & 10\% \\
        C-MET-CVPR'26   & 0\% & 5\% & 0\%  \\
        EmoTag-CVPR'26  & 15\% & 30\% & 10\%  \\
        \textbf{Ours}   & \textbf{70\%} & \textbf{60\%} & \textbf{65\%}  \\
        \bottomrule
    \end{tabular}
    }
    \caption{
    User study on emotional expressiveness, lip synchronization, and visual realism.
    }
    \label{tab:user_study}
\end{table}

\begin{figure}[ht!]
  \centering
    \includegraphics[width=0.9\linewidth]{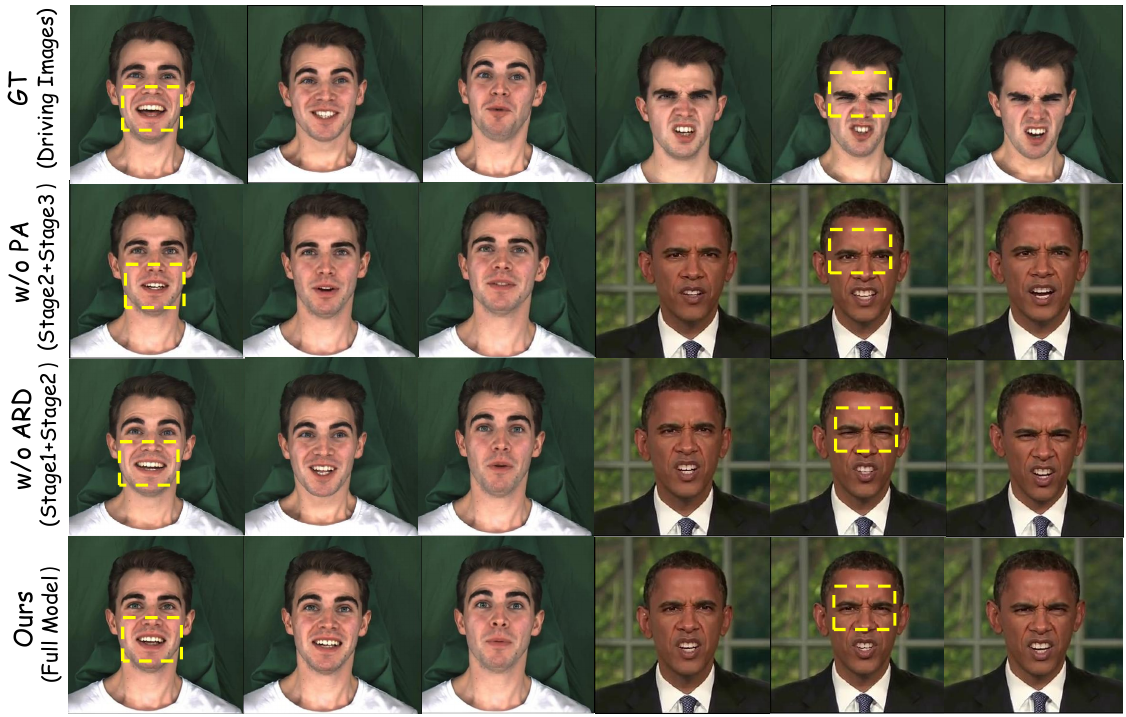}
\caption{
Qualitative ablation results. The left part shows results under the self-reconstruction setting, and the right part shows results under the cross-identity driving setting.
}
    \label{fig:ablation_proxy_avatar and adr}
\end{figure}

\noindent\textbf{Ablation study.} To evaluate the contribution of each component, we ablate
by removing key modules of our method.

\noindent\textbf{Proxy Avatar (PA).}
Fig. \ref{fig:ablation_proxy_avatar and adr} shows the qualitative ablation of the emotion proxy avatar. In the self-reconstruction setting, the full model produces results that are closer to the ground truth, especially in emotion-related facial regions such as the eyes, cheeks, and mouth. Without the emotion proxy avatar, the generated expressions become weaker and less consistent with the target emotional motion. In the cross-identity driving setting, the full model also exhibits stronger and more vivid emotional expressions, demonstrating that the proxy avatar provides effective emotion-aware motion priors for both reconstruction and cross-identity animation. Tab. \ref{tab:self-recon-tab}  and Tab. \ref{tab:cross_identity_quantitative}  report the quantitative ablation of the emotion proxy avatar. Removing the proxy avatar leads to consistent performance degradation in both self-reconstruction and cross-identity driving settings. These results demonstrate that the proxy avatar provides effective emotion-aware motion priors for expressive and identity-preserving portrait animation. The FPS remains almost unchanged because the final portrait generation stage always requires a driving video as input, regardless of whether it is produced by the emotion proxy avatar or obtained from an alternative source.

\noindent\textbf{Appearance Redundancy Distillation (ARD).} Fig. \ref{fig:ablation_proxy_avatar and adr} shows the qualitative ablation of ARD (Since the qualitative results of w/o and w/ few-step are difficult to distinguish by the naked eye, they are not presented). As highlighted by the yellow dashed boxes, introducing the acceleration module may cause slight appearance structure degradation in some local regions, such as subtle facial contours or fine texture details. However, these artifacts remain visually acceptable and do not significantly affect identity perception, emotional expression, or lip synchronization. Considering the substantial improvement in inference speed, the proposed module achieves a favorable balance between visual fidelity and real-time efficiency.
Tab.~\ref{tab:self-recon-tab} and Tab.~\ref{tab:cross_identity_quantitative} report the ablation results of ARD (w/o LRC(Low-Rank Caching) + w/o few-step and w/o LRC + w/ few-step). For the acceleration-related variants, the model without LRC and few-step distillation achieves the best numerical scores, which is expected because it uses the original multi-step denoising process and full reference appearance computation. However, the improvements over our full model are marginal, e.g., only 0.23 in FID and 0.045 in Lip Sync, while its inference speed is much lower as shown in Tab.~\ref{tab:self-recon-tab}. The variant with few-step distillation but without LRC also obtains results very close to our full model, indicating that low-rank appearance caching introduces only minor quality variations. Overall, these results show that our full model preserves comparable cross-identity generation quality while achieving substantially better inference efficiency.

\noindent\textbf{Other experimental analyses}. 
Due to space limitations, we provide more experimental analyses in the \textbf{appendix}. In particular, we analyze whether the single-identity proxy avatar introduces actor-specific motion bias and whether its emotional motion prior generalizes to diverse target portraits. We also quantify the storage overhead of the learnable low-rank matrices and evaluate the generalization of appearance caching across different identities and backgrounds.

\section{Conclusion}
In this paper, we propose a real-time one-shot emotion-controllable portrait animation framework. Our method first builds a 3D Gaussian-based emotion proxy avatar to generate expressive and controllable driving motions, and then transfers the identity-independent motion to arbitrary target portraits through a one-shot portrait driving module. To improve inference efficiency, we further introduce appearance redundancy distillation, which reduces redundant appearance computation while preserving visual fidelity. Extensive experiments demonstrate that our method achieves superior emotional expressiveness, identity preservation, lip synchronization, and real-time performance. We discuss limitations and potential future directions in the \textbf{appendix}.
\bibliography{aaai2027}

%\input{ReproducibilityChecklist.tex}

\end{document}